\documentclass[times]{elsarticle}

\usepackage{bsymb}

\usepackage{cite}
\usepackage{amsmath,amssymb,amsfonts}
\usepackage{relsize}
\usepackage{algorithmic}
\usepackage{graphicx}
\usepackage{subfig}

\usepackage{textcomp}
\def\BibTeX{{\rm B\kern-.05em{\sc i\kern-.025em b}\kern-.08em
    T\kern-.1667em\lower.7ex\hbox{E}\kern-.125emX}}

\usepackage{mathtools}

\usepackage[utf8]{inputenc}
\usepackage[T1]{fontenc}
\usepackage{microtype}

\usepackage{color}
\definecolor{gray}{rgb}{0.4,0.4,0.4}
\definecolor{darkblue}{rgb}{0.0,0.0,0.6}
\definecolor{cyan}{rgb}{0.0,0.6,0.6}
\definecolor{keycolor}{rgb}{0,0,0.8}     
\definecolor{labelcolor}{rgb}{0,0.4,0.8} 
\definecolor{codecolor}{rgb}{0,0,0}      
\definecolor{inhcolor}{rgb}{0.6,0.2,0}   
\definecolor{cmtcolor}{rgb}{0,0.4,0}     

\usepackage{bsymb}

\usepackage{listings}

\usepackage{color}
\definecolor{gray}{rgb}{0.4,0.4,0.4}
\definecolor{darkblue}{rgb}{0.0,0.0,1.0}
\definecolor{cyan}{rgb}{0.0,0.6,0.6}

\lstdefinelanguage{MPS}
{
  morestring=[b]",
  morestring=[s]
  ,
  morecomment=[s]{/*}{*/},
  stringstyle=\color{black},
  identifierstyle=\color{black},
  keywordstyle=\color{darkblue},
  morekeywords={domain,model,parent,concepts,concept,is,variable,
  individuals,deduced,attribute,domain,dom,relations,relation,attributes,
  enumerated,elements,
  range,functional,total,maplets,custom,data,sets,set,values,value,type,lexical,form,
  predicates,p1,p2,not}
  ,
  otherkeywords = {=,&,(,),\{,\},>,<,",:,?},
}

\usepackage{lineno,hyperref}
\modulolinenumbers[5]

\journal{Arxiv}

\usepackage{breqn}
\usepackage{longtable}

\usepackage[linewidth=1pt]{mdframed}
\usepackage{multicol}

\usepackage{flushend}

\begin{document}

\begin{frontmatter}

\title{Towards Linearization Machine Learning Algorithms
}


\author{Steve Tueno}
\address{Université de Sherbrooke, Québec, Canada}
\ead{steve.jeffrey.tueno.fotso@usherbrooke.ca}


\begin{abstract}
This paper is about  a machine learning approach based on the multilinear projection of an unknown function (or probability distribution) to be estimated towards a linear (or multilinear) dimensional space \texttt{E'}.
The proposal transforms the problem of predicting the target of an observation \texttt{x} into a problem of determining a consensus among the \textit{k} nearest neighbors of  \texttt{x}'s image within the dimensional space \texttt{E'}.
The algorithms that concretize it allow both  regression and  binary classification.
Implementations carried out using \textit{ Scala/Spark} and assessed on a dozen \textit{LIBSVM} datasets have demonstrated  improvements in prediction  accuracies  in comparison with other prediction algorithms implemented within \textit{Spark MLLib} such as  multilayer perceptrons, logistic regression classifiers and random forests.


\end{abstract}


\end{frontmatter}


\section*{Introduction}
This work focuses on the  supervised learning problem: exploitation of historical data for classification (prediction of discrete values) \citep{breiman2017classification} and regression (prediction of continous values) purposes \citep{quinlan1992learning}.


Several approaches are proposed to solve the supervised learning problem like for instance:

\begin{itemize}
\item[•] \textbf{Linear regression} \citep{seber2012linear} which consists in using an equation such as  $Y=W^{t}X$ to predict the output $Y$ that corresponds to an observation $X = (1, x_1, ..., x_n)$. Vector  $W= (w_0, w_1, ..., w_n)$ represents the parameters estimated from historical data $(X_0, Y_0)$: $W = (X_{0}^{t}X_{0})^{-1}X_{0}^{t}Y_{0}$ (least squares estimate).

\item[•]  \textbf{Logistic regression} \citep{kleinbaum2002logistic} which is widely used to predict a binary response and consists in using an equation such as  $p=1/(1+e^{-(W^{t}X)})$ to predict the output $y$ that corresponds to an observation $X = (1, x_1, ..., x_n)$: here, $y$ is a binary class whose value depends on whether  $p>0.5$. 

\item[•]  \textbf{Random forests} \citep{breiman2001random} which combine separately trained decision trees  \citep{rokach2008data} to improve prediction accuracy. Randomness is injected when training decision trees to reduce the risk of overfitting. Each prediction is obtained through consensus (aka  aggregating the results) from the combined decision trees: majority vote in case of classification and averaging in case of regression.

\item[•]  \textbf{Multilayer perceptron} \citep{pal1992multilayer} which consists of several layers of nodes forming a network, each layer being fully connected to the next layer in the network. Nodes in the input layer represent the input data while the other nodes represent  linear combinations acting on inputs associated with an activation function. Nodes in the output layer are used to produce the prediction and their number corresponds to the number of classes (number of possible output values).
\end{itemize}

This paper introduces  a novel machine learning approach based on the multilinear projection of an unknown function (or probability distribution) to be estimated towards a linear (or multilinear) dimensional space \texttt{E'}. This transforms the supervised learning problem into a problem of determining a consensus among the \textit{k} nearest neighbors of  image of an input observation \texttt{x}, within the dimensional space \texttt{E'}.
The proposal is concretized into algorithms that  allow both  regression and  binary classification. For a multiclass classification, one can consider  transformation to binary multi-class classification techniques such as \textit{one-against-all} \citep{bishop2006pattern}.
Implementations \citep{scala_implementation_linearization_ml_approach} carried out using \textit{ Scala/Spark} \citep{odersky2008programming,pentreath2015machine} and assessed on a dozen \textit{LIBSVM} datasets (\textit{breast-cancer, a1a, a2a, iris_libsvm, square root function dataset\footnote{A dataset that simulates the square root function using a thousand randomly picked points.}, etc.} \citep{zeng2008fast,libsvmtools_datasets_link}  have demonstrated  improvements in prediction  accuracies  in comparison with \textit{Spark MLLib} \citep{meng2016mllib} implementations of   multilayer perceptrons, logistic regression classifiers and random forests.

In the following, we present the approach, describe the algorithms and discuss the assessments performed.


\section{The Linearization Machine Learning Approach}

\subsection{Overview}

 \begin{figure*}

\begin{center}
\includegraphics[width=1.\linewidth]{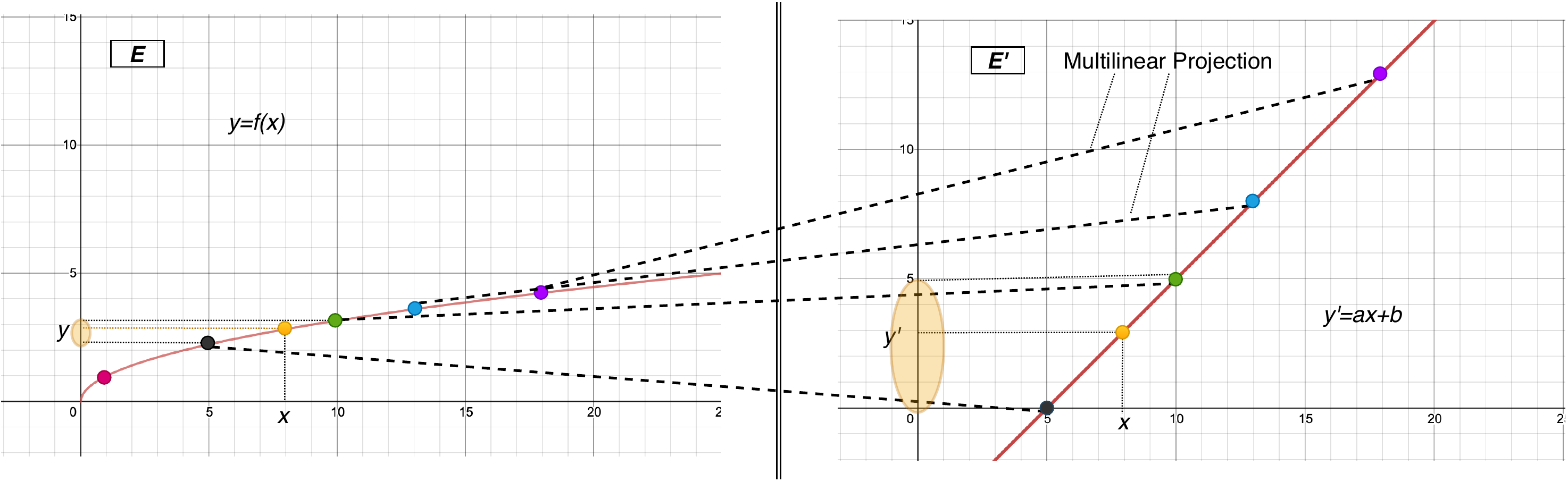}\caption{\label{Linearization_Machine_Learning_Approach_Overview} Overview of the linearization machine learning approach} 
\end{center}
 
\end{figure*}

 Figure \ref{Linearization_Machine_Learning_Approach_Overview}  provides an overview of the linearization machine learning approach. Its Scala implementation is available at \citep{scala_implementation_linearization_ml_approach}.
 An unknown function (or probability distribution) $f(x)$ (respectively $f(X)$ with $X = (1, x_1, ..., x_n)$) is projected within a  space \textit{E'} into a linear (respectively multilinear) function $f'(x) = ax+b$ (respectively $f'(X)=W^{t}X$ with $W= (w_0, w_1, ..., w_n)$). The multilinear projection consists of linear transformations of points of the historical dataset (or learning dataset): each linear transformation is defined to map a point $(x_i, y_i=f(x_i))$ of $E$ into a point $(x_i, y'_i=f'(x_i))$ of $E'$.
 Coefficients $(a, b) or (w_0, w_1, ..., w_n)$ are parameters of the learning model. They must be fine tuned using  optimization algorithms.

Let $x$ be  a new observation and $(y'_{j_1}, ...y'_{j_k})$ be the \textit{k} nearest neighbors of $y'=f'(x)$ following a distance such as the absolute value of the difference ($k \in 1..n$ is also a parameter of the learning model which must be fine tuned). The prediction $y$ is given by:

\begin{equation*}
y = 
\frac{1}{k}
\mathlarger{\mathlarger{\sum}}_{t=j_1}^{j_k}
\frac{y' * y_{j_t}}{y'_{j_t}}
\end{equation*}

This estimate is obtained by average. The median  can also be considered.

\subsection{Algorithms}

\subsubsection{Regression}
\begin{enumerate}
\item Estimate the best parameter values using an optimization algorithm
\item For an input $X$:
\begin{enumerate}
\item Compute the subset $(y'_{j_1}, ...y'_{j_k})$ of the \textit{k} nearest neighbors of $y'=f'(X)$
\item output 
$
y := 
\frac{1}{k}
\mathlarger{\mathlarger{\sum}}_{t=j_1}^{j_k}
\frac{y' * y_{j_t}}{y'_{j_t}}
$
\end{enumerate}
\end{enumerate}

\subsubsection{Binary Classification}
Let 0 and 1 be the labels and $random$ be a function that allows  to randomly choose a value in a given set.

\begin{enumerate}
\item For each historical data point $(X_i, c_i)$ where $i \in 1..n$:
\begin{enumerate}
\item If $c_i=0$, set $p_i := random(]0, 0.5[)$

Else, set $p_i := random(]0.5, 1[)$
\end{enumerate}

\item Set $(y_i)_{i \in 1..n} := Learn((X_i)_{i \in 1..n}, (c_i)_{i \in 1..n}, (p_i)_{i \in 1..n})$

\item For an input $x$:
\begin{enumerate}
\item Compute the subset $(y'_{j_1}, ...y'_{j_k})$ of the \textit{k} nearest neighbors of $y'=f'(x)$
\item If
$ 
\frac{1}{k}
\mathlarger{\mathlarger{\sum}}_{t=j_1}^{j_k}
\frac{y' * y_{j_t}}{y'_{j_t}}
 > 0.5$, output 1
 
 Else, output 0
\end{enumerate}
\end{enumerate}

\subsubsection{Function Learn}

\begin{enumerate}
\item For each historical data point $(X_i, c_i)$ where $i \in 1..n$:
\begin{enumerate}
\item Compute the subset $(y'_{j_1}, ...y'_{j_k})$ of the \textit{k} nearest neighbors of $y'_i=f'(X_i)$
\item Set $
q_i := 
\frac{1}{k}
\mathlarger{\mathlarger{\sum}}_{t=j_1}^{j_k}
\frac{y'_i * p_{j_t}}{y'_{j_t}}
$
\item If $|q_i - p_i|>>0$

\begin{enumerate}
\item  If $((c_i = 0 \wedge q_i < 0.5) \vee (c_i = 1 \wedge q_i \in  [0.5,1]))$, set $p_i := q_i$

      Else if $(q_i  >> p_i  \wedge  ((c_i = 0 \wedge  p_i + inc < 0.5) \vee (c_i = 1 \wedge p_i + pas \leq 1)))$  , set $p_i := p_i+inc$

      Else if $(q_i   << p_i \wedge ((c_i = 0 \wedge p_i - inc \geq 0) \vee (c_i = 1 \wedge p_i - inc \geq 0.5)))$, set $p_i := p_i-inc$
\end{enumerate}
\end{enumerate}

\item If $(p_i)_{i \in 1..n}$ has not changed, output $(p_i)_{i \in 1..n}$

Else output $Learn((X_i)_{i \in 1..n}, (c_i)_{i \in 1..n}, (p_i)_{i \in 1..n})$

\end{enumerate}

\section{Discussion}
The datasets and approach  implementations are available at \citep{scala_implementation_linearization_ml_approach}.
Our experience, without proper parameter tuning,  has been that the  linearization machine learning implementations are in many cases  more accurate and must be further investigated.

\begin{table}[htb]
\centering
\caption{\label{assessment_overview} Overview of assessment accuracies}
\begin{scriptsize}
\begin{tabular}{|p{1.5cm}|p{3.5cm}|p{3.cm}|p{3.cm}|p{3.cm}|}
\hline
\textbf{Dataset} & \textbf{ Multilayer perceptrons} & \textbf{Logistic regression}  & \textbf{Linearization ML Approach}\\
\hline
\textbf{\textit{breast-cancer}} & 96\% (238/248) & 86\% (213/248)  & 98\% (241/248)   \\
\hline
\textbf{\textit{a1a}} & 79\% (459/582) & 75\% (436/582) & 74\% (425/582) \\
\hline
\textbf{\textit{square root}} & 73\% (2560/3507) & 57\% (1999/3507)  & 90\% (3147/3507)  \\
\hline
\textbf{\textit{exp}} & 77\% (3087/4010) & 77\% (3087/4010)  & 78\% (3107/4010)  \\
\hline
\textbf{\textit{cod-rna}} & 93\% (19463/20929) & 66\% (13813/20929)  & 86\% (17790/20929)  \\
\hline
\end{tabular}
\end{scriptsize}
\end{table}

\section*{References}



\bibliographystyle{elsarticle-num}
\bibliography{references}

\end{document}